\documentclass[11pt, conference, compsocconf]{IEEEtran}
\ifCLASSINFOpdf
\else
\fi

\usepackage{color, todonotes}
\usepackage{subfigure}
\usepackage{algorithm}
\usepackage{algorithmic}
\usepackage{amsmath}
\usepackage{amssymb}
\usepackage{graphicx}
\usepackage{multirow}
\usepackage{url}

\newcommand{\R}{\mathbb{R}}
\DeclareMathOperator*{\argmin}{argmin}

\begin{document}
%
\title{Optimizing the Union of Intersections LASSO ($UoI_{LASSO}$) and Vector Autoregressive ($UoI_{VAR}$) Algorithms for Improved Statistical Estimation at Scale} 


\author{

\IEEEauthorblockN{Mahesh Balasubramanian\IEEEauthorrefmark{1}\IEEEauthorrefmark{3}, Trevor Ruiz\IEEEauthorrefmark{4}, Brandon Cook\IEEEauthorrefmark{3}, Sharmodeep Bhattacharyya\IEEEauthorrefmark{4}, \\
Prabhat\IEEEauthorrefmark{3}, Aviral Shrivastava\IEEEauthorrefmark{1}, Kristofer E. Bouchard\IEEEauthorrefmark{2}\IEEEauthorrefmark{5}
}

\IEEEauthorblockA{
\IEEEauthorrefmark{1}Compiler Microarchitecture Lab, Arizona State University, Tempe, AZ \\
\IEEEauthorrefmark{2}Biological Systems and Engineering Division, LBNL, Berkeley, CA. \\
\IEEEauthorrefmark{3}NERSC, Lawrence Berkeley National Laboratory, Berkeley, CA. \\
\IEEEauthorrefmark{4} Statistics Department, Oregon State University, OR. \\
\IEEEauthorrefmark{5} Senior Author\\
Email: \{mbalasubramanian\}@asu.edu, \{kebouchard\}@lbl.gov} }
\maketitle

\begin{abstract}
The analysis of scientific data of increasing size and complexity requires statistical machine learning methods that are both interpretable and predictive. Union of Intersections (UoI), a recently developed framework, is a two-step approach that separates model selection and model estimation. A linear regression algorithm based on UoI, $UoI_{LASSO}$, simultaneously achieves low false positives and low false negative feature selection as well as low bias and low variance estimates. Together, these qualities make the results both predictive and interpretable. In this paper, we optimize the $UoI_{LASSO}$ algorithm for single-node execution on NERSC's Cori Knights Landing, a Xeon Phi based supercomputer. We then scale $UoI_{LASSO}$ to execute on cores ranging from 68-278,528 cores on a range of dataset sizes demonstrating the weak and strong scaling of the implementation. We also implement a variant of $UoI_{LASSO}$, $UoI_{VAR}$ for vector autoregressive models, to analyze high dimensional time-series data. We perform single node optimization and multi-node scaling experiments for $UoI_{VAR}$ to demonstrate the effectiveness of the algorithm for weak and strong scaling. Our implementations enable to use estimate the largest VAR model (1000 nodes) we are aware of, and apply it to large neurophysiology data 192 nodes).

\end{abstract}



%
\maketitle

\section{Introduction}


The growth of the Internet and social media applications have paved the way for the development of highly sophisticated machine learning and statistical data analysis tools. Beyond the Internet, scientific data collection strategies have grown exponentially over the years by innovation in the field of sensors and advanced data collection methods. Many fields such as genetics,  mass spectrometry, and neuroscience now have the means of collecting big data through various devices and sensors~\cite{national2013frontiers}. In particular, advanced recording devices created as part of the BRAIN Initiative enable recording neural activities from hundreds to thousands of neurons for days at a time, generating TeraBytes and some cases PetaBytes of data~\cite{8352088}.  

Statistical data analysis methods for scientific applications are required to give comprehensible results from large datasets in a human-readable way (interpretable) while maintaining high-quality prediction (predictive). However, there is generally a trade-off between interpretability and predictability, methods offering both are uncommon \cite{national2013frontiers}. In particular, most commercial applications, which are the major forcing functions in stats/ML methods development, are comfortable with black-box approaches that maximize predictive accuracy on, e.g., predicting how many clicks an ad will get. This is particularly true for method designed to be run at large scale. Innovations in the supercomputing domain enable the analysis of large-scale datasets, especially if the data analysis framework is highly scalable (execute on 10000s of cores).

The Union of Intersections (UoI) framework developed in~\cite{bouchard2017union} is a novel statistical machine learning framework that enhances the interpretability and predictive accuracy of many methods while also being scalable to analyze big datasets. Methods based on UoI perform model selection and model estimation through intersection and union operations, respectively. For both selection and estimation, UoI-based methods utilize the notion of `stability to perturbation'. 
The main mathematical innovations of the $UoI$ algorithm are 1) create a family of potential model supports through an intersection operation for a range of regularization parameters; and 2) combine the above-computed supports with a union operation so as to increase prediction accuracy on held out data.

In this paper,  we implement $UoI_{LASSO}$ and $UoI_{VAR}$ in C++ and analyze the algorithms for optimizations for better performance on single-node of Xeon Phi processor supercomputer system. Then, we analyze the natural algorithmic parallelism and try to exploit it in our implementation using the \texttt{MPI} framework. We evaluate our multi-node implementation with high-dimensional synthetic datasets and demonstrate the weak and strong scaling of both the algorithms. Finally, we run the $UoI_{VAR}$ algorithm on real neurophysiology data to create a the largest (192 nodes) VAR model in neuroscience.

\section{Methods}
\subsection{Overview of UoI Algorithm}
The UoI is a generalized framework into which a wide range of existing algorithms can be implemented. In this paper, we implement $UoI_{LASSO}$ to solve the sparse linear regression problem, which is subsequently used to enable $UoI_{VAR}$ for vector autoregressive models to analyze high dimensional time series data. A serial version of the $UoI_{LASSO}$ algorithm is given in Algorithm~\ref{algoLass} and $UoI_{VAR}$ is given in algorithm~\ref{algoVAR}. The basic framework has two modules, namely model selection, and model estimation. In model selection, for a range of bootstrap subsamples ($B_{1}$) and regularization parameters ($\lambda$), the LASSO algorithm is used to find estimates by solving a constrained convex optimization problem. For a given value of ($\lambda$) LASSO algorithm sets some features exactly to zero on any given bootstrap sample. The support associated with a given $\lambda$, $S_{j}$ is taken as the intersection of the supports across bootstrap samples. This is done for each value of ($\lambda$), creating a family of potential model supports $\textbf{S} = [S_{1}, S_{2},...,S_{q}]$. In the model estimation module, estimates for the different model supports ($S_{j}$) are calculated with ordinary-least-squares (OLS), a nearly unbiased estimator, across a number of bootstrap samples $B_{2}$. The estimates associated with different model supports are averaged (a union operation) so as to increase prediction accuracy. This last step is a novel model averaging, merging some ideas of bagging (averaging parametric estimates) and boosting (combining results across multiple models).

\subsection{Formal Statistical Description}
Let us consider $n$ samples of input data ($(Y_{1}, X_{1})$, $...$,$(Y_{n}, X_{n})$) with univariate response variable $Y$ and $p$-dimensional predictor variable $X$. The linear regression model for this input data is generated as:

\begin{equation} \label{eq1}
Y = X\beta + \epsilon
\end{equation}

where $Y$ $=$ $(Y_{1},...,Y_{n}),X$ is a $n\times\/p$  design matrix; $\epsilon$ $=$ $(\epsilon_{1},...,\epsilon_{n})$ are random noise terms with $\epsilon\overset{iid}{\sim}N(0, \sigma^2$\textbf{I}$_n)$. Let $S$ be a non-zero coefficient set of $\beta$.

The LASSO regression algorithm with penalization parameter $\lambda >  0$ minimizes the following constrained convex optimization problem with respect to $\beta$:
\begin{equation} \label{eq2}
\hat{\beta} = argmin_{\beta} ||Y - X\beta||^2 +  \lambda||\beta||_1
\end{equation}
Here, the first term on the right-hand side penalizes the error of the estimates, while the second term penalizes $L_{1}$ norm of the parameter vector $\beta$, setting some values of $\beta$ to zero.


\subsection{Model Selection -- Intersection}
For every bootstrap sample $T^k$ the LASSO estimates ($^j\hat{\beta}^{k}$) are computed (here, using the LASSO Alternating Direction Method of Multipliers (ADMM), see equation~\ref{eq5}) across different regularization parameter values, $\lambda_j$. For each bootstrap sample, the support ($S_j^k$) are the non-zero values of the estimates calculated by LASSO-ADMM.

It is known that the LASSO estimator is prone to false positives: i.e., it includes more parameters than are in the model. To mitigate this, the support associated with a given $\lambda$, $S_{j}$ is taken as the intersection of the supports across bootstrap samples: 

\begin{equation} \label{intersection}
{S}_j = \bigcap\limits_{k=1}^{B_1} S_j^k
\end{equation}

This is done for each value of ($\lambda$), creating a family of potential model supports $\textbf{S} = [S_{1}, S_{2},...,S_{q}]$.

\subsection{Model Estimation -- Union}
A $B_2$ number of estimation bootstraps are considered to compute the model estimates. A low bias estimator, like Ordinary Least Squares (OLS), is used to estimate the associated model from the model selection step (Algo.\ref{algoLass} line 18). The algorithm then computes a \textit{Union} of supports by averaging OLS estimates.

The variable set post union (averaging) can be represented as (approximately): 

\begin{equation} \label{union}
S_{UoI} = \bigcup\limits_{l=1}^{B_2} S_{jl} = \bigcup\limits_{l=1}^{B_2} \bigcap\limits_{k=1}^{B_1}S_{jl}^k
\end{equation}

\subsection{Distributed Constrained Convex Optimization by Alternating Direction Method of Multiplier}
Here, we use the Alternating Direction Method Multiplier (ADMM) to minimize the $L_1$ regularized linear regression (Equation~\ref{eq2}). LASSO-ADMM solves the dual problem in form of equation \ref{eq_ADMM}: 

\begin{equation} \label{eq_ADMM}
\begin{split}
\textnormal{minimize } f(x) + g(z) \\
\textnormal{subject to } x - z = 0 \\
\textnormal{where, }  f(x) = (1/2) ||Y - X\beta||^2_{2}; \\
g(z) = \lambda||\beta||_1
\end{split}
\end{equation}

where x $\in \R^n$ , z $\in \R^m$, and $f$ and $g$ are convex. The LASSO-ADMM algorithm consists of an $x$ minimization, $z$ minimization followed by a dual variable update. The separation of minimization over $x$ and $z$  allows for the separate decomposition of $f$ or $g$. Here, $x$ and $z$ can be updated in sequential or alternating computations which gives the name alternating direction. In the distributed ADMM algorithm, each core is responsible for computation of its own objective and constraint variables and its quadratic term is updated so that all the cores converge to a common value of estimates. To ensure a good scalability, the ordinary least squares (OLS) is implemented using LASSO-ADMM algorithm for model estimation by setting regularization parameter $\lambda$ to 0, thereby making $g$ in equation~\ref{eq_ADMM} equal to 0.

\begin{algorithm}[t!]
\caption{$UoI_{LASSO}$ $(Input Data(X,y) \in \R^{n\times(p+1)},
\lambda \in \R^{q}$, $B_{1}$, $B_{2}$)}
\begin{algorithmic}[1] 
	\STATE \textbf{Model Selection}
    \FOR{$k=1$ to $B_{1}$}
		\STATE Generate bootstrap sample $T^{k}$ $=$ ($X^{k}_{T}, Y^{k}_{T}$)
        \FOR{$\lambda_j$ $\in$ $\lambda$}
   	 		\STATE Compute LASSO estimate $^j\hat{\beta}^{k}$ from $T^{k}$
            \STATE Compute support $S^{k}_{j}$ $=$ $\{i\}$ s.t $^j\hat{\beta}^{k}_{i}$ $\neq$ 0
        \ENDFOR
    \ENDFOR
    \FOR{$j=1$ to $q$}
    	\STATE Compute Bootstrap-LASSO support \newline for $\lambda_{j}$ : $S_{j}$ $=$ $\bigcap\limits_{k=1}^{B_{1}}$ $S_{j}^{k}$ (as in equation~\ref{intersection})
    \ENDFOR
    \STATE \textbf{Model Estimation}
    	\FOR{$k=1$ to $B_2$}
        	\STATE Generate bootstrap samples for training and evaluation:
            \STATE training $T^{k}$ $=$ ($X^{k}_{T}, Y^{k}_{T}$)
            \STATE evaluation $E^{k}$ $=$ ($X^{k}_{E}, Y^{k}_{E}$)
            \FOR{$j=1$ to $q$}
            	\STATE Compute OLS estimate $\hat{\beta_{S_{j}}^{k}}$ from $T^{k}$
                \STATE Compute loss on $E^{k}$ : $L(\hat{\beta_{S_{j}}^{k}},E^{k} )$
            \ENDFOR
            	\STATE Compute best model for each bootstrap sample:	
                \STATE $\hat{\beta_{S}^{k}}$ $=$ $\argmin\limits_{\hat{\beta_{S_{j}}^{k}}}$ $L(\hat{\beta_{S_{j}}^{k}},E^{k} )$
        \ENDFOR
        \STATE Compute averaged model estimates $\hat{\beta^{*}}$ $=$ $\frac{1}{B_{2}}$ $\sum\limits_{k=1}^{B_{2}}$\!$\hat{\beta_{S}^{k}}$ (as in equation~\ref{union})
        \STATE Return: $\hat{\beta^{*}}$
\end{algorithmic}
\label{algoLass}
\end{algorithm}

\subsection{$UoI_{LASSO}$ Algorithm}
A high-level overview of the $UoI_{LASSO}$ algorithm consists of two Map-Solve-Reduce steps (Figure 1). The algorithm takes multiple random bootstrap subsamples of the input data ($Map$) and distributes it across different computation cores. Next, LASSO and OLS ($Solve$) use the data for solving the convex optimization. The resultant estimates are then combined by intersection and union operations ($Reduce$). The $Reduce$ step in model selection performs a feature compression by intersection operation of supports across bootstraps. The $Reduce$ step in model estimation performs a feature expansion by averaging (union operation) the OLS estimates across potentially different model supports. 


\begin{algorithm}[!ht!]
\caption{$UoI_{VAR}$ $(Input Data (X_1, \dots, X_N)^T \in \R^{N\times p)}, \lambda \in \R^{q}$, $B_{1}$, $B_{2}$)}
\begin{algorithmic}[1] 
	\STATE \textbf{Model Selection}
    \FOR{$k=1$ to $B_{1}$}
		\STATE Generate bootstrap sample $T^{k}$ $=$ ($X^{k}_{T1}, \dots, X^{k}_{TN}$)
        \STATE Construct $(\mathbf{Y}^k_T, \mathbf{X}^k_T)$ (as in equations~\ref{eq6} - ~\ref{eq7})
        \STATE Construct $Y^k_T = \text{vec}\mathbf{Y}^k_T$ and $X^k_T = (\mathbf{I}\otimes\mathbf{X}^k_T)$
        \FOR{$\lambda_j$ $\in$ $\lambda$}
   	 		\STATE Compute LASSO estimate $^j\hat{\beta}^{k}$ from $(X^{k}_T, Y^k_T)$
            \STATE Compute support $S^{k}_{j}$ $=$ $\{i\}$ s.t $^j\hat{\beta}^{k}_{j}$ $\neq$ 0
        \ENDFOR
    \ENDFOR
    \FOR{$j=1$ to $q$}
    	\STATE Compute Bootstrap-LASSO support \newline for $\lambda_{j}$ : $S_{j}$ $=$ $\bigcap\limits_{k=1}^{B_{1}}$ $S_{j}^{k}$ (as in equation~\ref{intersection})
    \ENDFOR
    \STATE \textbf{Model Estimation}
    	\FOR{$k=1$ to $B_2$}
        	\STATE Generate bootstrap samples for training and evaluation:
            \STATE training $T^{k}$ $=$ ($X^{k}_{T1}, \dots, X^{k}_{TN}$)
            \STATE evaluation $E^{k}$ $=$ ($X^{k}_{E1}, \dots, X^{k}_{EN}$)
            \STATE Construct $(\mathbf{Y}^k_T, \mathbf{X}^k_T)$ (as in equations~\ref{eq6} - ~\ref{eq7})
            \STATE Construct $(\mathbf{Y}^k_E, \mathbf{X}^k_E)$ (as in equations~\ref{eq6} - ~\ref{eq7})
        	\STATE Construct $Y^k_T = \text{vec}\mathbf{Y}^k_T$ and $X^k_T = (\mathbf{I}\otimes\mathbf{X}^k_T)$
            \STATE Construct $Y^k_E = \text{vec}\mathbf{Y}^k_E$ and $X^k_E = (\mathbf{I}\otimes\mathbf{X}^k_E)$
            
        \FOR{$j=1$ to $q$}
            	\STATE Compute OLS estimate $\hat{\beta_{S_{j}}^{k}}$ from $T^{k}$
                \STATE Compute loss on $E^{k}$ : $L(\hat{\beta_{S_{j}}^{k}},E^{k} )$
            \ENDFOR
            	\STATE Compute best model for each bootstrap sample:	
                \STATE $\hat{\beta_{S}^{k}}$ $=$ $\argmin\limits_{\hat{\beta_{S_{j}}^{k}}}$ $L(\hat{\beta_{S_{j}}^{k}},E^{k} )$
        \ENDFOR
        \STATE Compute averaged model estimates \\$\hat{\beta^{*}}$ $=$ $\frac{1}{B_{2}}$ $\sum\limits_{k=1}^{B_{2}}$\!$\hat{\beta_{S}^{k}}$ (as in equation~\ref{union})
        \STATE Partition $\hat{\beta}^*$ and rearrange into $(\hat{A}_1, \dots, \hat{A}_d)$ and $\hat{\mu}$
        \STATE Return: $(\hat{A}_1, \dots, \hat{A}_d)$ and $\hat{\mu}$
\end{algorithmic}
\label{algoVAR}
\end{algorithm}

\subsection{$UoI_{VAR}$ Algorithm}
The $UoI_{LASSO}$ implementation can be adapted with small modifications to sparse estimation of vector autoregressive model parameters from high-dimensional time series data. In this case the input data is a vector time series $\{X_t\}_{t = 1}^N$ generated by a vector autoregressive process of order $d$, $VAR(d)$:
\begin{equation}
X_t = \sum_{j = 1}^d A_j X_{t - j} + U_t
\label{eq5}
\end{equation}
where $X_t\in\mathbb{R}^p$, and the process has $p$-dimensional Gaussian disturbances $U_t \stackrel{iid}{\sim} \mathbb{N}_p (0, \Sigma)$. The stability of the process is expressed by the constraint $\textbf{det}(I - \sum_{j = 1}^d A_j z^j) \neq 0 \quad\forall\; |z| \leq 1$.

Equation~\ref{eq5} provides a model for the data which can be written as a multivariate least squares problem with correlated errors of the form $\mathbf{Y} = \mathbf{XB} + \mathbf{E}$. In particular, the response is the $(N - d) \times p$ matrix 
\begin{equation}
\mathbf{Y} = (X_N, X_{N - 1}, \dots, X_{d + 1})^T
\label{eq6}
\end{equation}
and the regressors are lagged values represented in the $(N - d)\times(dp)$ matrix 
\begin{equation}
\mathbf{X} = \left(\begin{array}{cccc}
	X_{N - 1}' &X_{N - 2}' &\dots &X_{N - d}' \\
    X_{N - 2}' &X_{N - 3}' &\dots &X_{N - (d + 1)}' \\
    \vdots &\vdots &\ddots &\vdots \\
    X_{d}' &X_{d - 1}' &\dots &X_1'
    \end{array}\right)
    \label{eq7}
\end{equation}
and the coefficient matrix is $\mathbf{B}' = (A_1 A_2 \dots A_d)$. Several estimation strategies are classically applied to this form of time series model for low-dimensional data (small $p$), among which is to vectorize the problem as shown in equation~\ref{eq8} and apply ordinary least squares to estimate the entries of the $A_j$ matrices. 

\begin{equation}
\text{vec}\mathbf{Y} = \left(\mathbf{I} \otimes \mathbf{X}\right) \text{vec}\mathbf{B} + \text{vec}\mathbf{E}
\label{eq8}
\end{equation}

Equation~\ref{eq8} then has the same form as equation~\ref{eq1}. Noting this correspondence, estimation with sparsity in high-dimensional time series can be accomplished by first rearranging the multivariate least squares problem and then solving the LASSO problem (equation~\ref{eq2}) for the resulting rearrangement.

The UoI implementation, shown as Algorithm \ref{algoVAR}, is consequently similar to $UoI_{LASSO}$, but with a bootstrap method appropriate for capturing temporal dependence in the input data and large matrix operations required to obtain a problem of the form shown in equation~\ref{eq2}. Aside from these modifications, the algorithm is the same as $UoI_{LASSO}$ Algorithm~\ref{algoLass}.


\begin{figure*}[t]
\centering
\includegraphics[width=\linewidth]{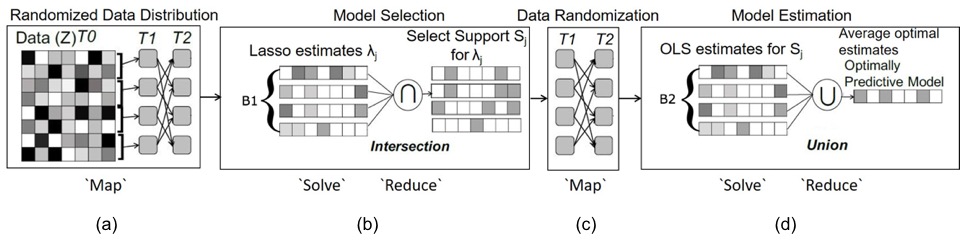}
\caption{(a) A Three-Tier (T0, T1 and T2) distribution strategy for randomized distribution of dataset across the number of sample from the HDF5 data file to the cores of KNL. (b) Model Selection -- LASSO ADMM is used to `Solve' and \textbf{Intersection} operation is used as `Reduce' to select family of support S$_{j}$. (c) Data randomization for cross validation where Tier2 random distribution is employed to randomly reshuffle the data. (d) Model Estimation -- OLS is used to `Solve' and \textbf{Union} operation is used to `Reduce' to get an optimally predictive model. 
}
\vspace{-0.3cm}
\label{data-distribution}
\end{figure*}

\section{Scalable $UoI_{LASSO}$ and $UoI_{VAR}$}
$UoI_{LASSO}$ and $UoI_{VAR}$ exhibits a high degree of algorithmic parallelism. In each of the model selection and model estimation steps, the bootstrap subsamples ($B_1$ and $B_2$) can be parallelized, referred to as \textit{P$_B$} parallelization. Additionally, parallelization over regularization parameters ($\lambda_j$) can be used (referred to as \textit{P$_\lambda$} parallelization). An important point to consider is that the model selection and model estimation must occur in a sequential order and cannot be parallelized. In addition to the \textit{P$_B$} and \textit{P$_\lambda$} parallelization, the LASSO-ADMM computation can be performed in a distributed manner. 


\subsection{Randomized Data Distribution Design for HDF5}
\subsubsection{$UoI_{LASSO}$}
The synthetic datasets used in this evaluation have the "Samples" in Rows and "Features" in Columns. The dataset size is the problem size for $UoI_{LASSO}$. We use \texttt{HDF5} application program interface for Data I/O. \texttt{HDF5} offers parallel reading of the input file, albeit in contiguous chunks. The library does not provide a random reading of input data without reading the file multiple times in a loop. To parallelize this operation, we introduce a novel randomized data distribution technique. First, the data is read in parallel from the input file into the computation cores in contiguous blocks. As shown in Figure~\ref{data-distribution} T0 or \textit{Tier0} is the source HDF5 file. The contiguous reading by all the processes is done in T1, \textit{Tier1}, using HDF5 hyperslabs~\cite{folk2011overview}. \textit{Tier0} and \textit{Tier1} data distribution use an underlying HDF5-parallel library for parallel accesses and hyperslab creation. By creating hyperslabs, the application can read the data file and load them into the memory space created on each compute core. Having loaded the data from the input file into the compute cores, we employ MPI One-Sided communication to randomly distribute the subsamples (T2, \textit{Tier2}).
\subsubsection{$UoI_{VAR}$}
In $UoI_{VAR}$ the input data is a time series and a temporal dependence is required while analyzing the data. To maintain this dependence, a block shuffle approach was adopted by randomly selecting time series blocks for every bootstrap subsample. The Algorithm~\ref{algoVAR}  lines 5 and 21-22, requires a column stacking vectorization step to construct $Y^k_T$ and an identity Kronecker product step to construct $X^k_T$. In the serial version of the algorithm a simple vectorization and Kronecker product functions can be invoked, but in a distributed-memory parallel paradigm, this is not possible. Unlike $UoI_{LASSO}$, the synthetic datasets for $UoI_{VAR}$ are relatively small (in order of MegaBytes) and the problem is created in the \textit{Kronecker Product} and \textit{Vectorization} (lines 5) of Algorithm~\ref{algoVAR}. The actual problem size increases in the order $\approx O(p^3)$, where $p$ is the number of features. Due to the small size of the data, the T1 parallel reading layer cannot be deployed. To overcome this issue, we have developed a distributed Kronecker product and vectorization strategy using \texttt{MPI} one-sided communication with the windows created by the \textit{n\_reader} processes: a small number of processes (usually equal to the number of samples based on the availability of resources) read the data file in parallel and creates windows for MPI-One sided communication for distributed Kronecker Product and Vectorization.

\section{Results}
The single node and multi-node runs for this paper were conducted on Cori Knights Landing (KNL) supercomputer at NERSC. Cori KNL is a Cray XC40 supercomputer consisting of 9,688 nodes of 1.4 GHz Intel Xeon Phi processors with a single socket 68 cores per node.   The aggregated memory for a single node in KNL is 16GB MCDRAM and a 96GB DDR.  The $UoI_{LASSO}$ and $UoI_{VAR}$ algorithms were implemented in C++ using Eigen3 library~\cite{eigenweb} for linear algebra computations and Intel-MKL library~\cite{wang2014intel} for BLAS operations for $UoI_{LASSO}$ to utilize the inbuilt Single Instruction Multiple Data (\texttt{SIMD}) directives. The \texttt{MPI} framework was used for parallelization and communication between the processes.
 
For all the evaluations in this paper, synthetic datasets ranging from 16GB to 8TB were generated for $UoI_{LASSO}$, and datasets that generate problem sizes of 16GB to 8TB were generated for $UoI_{VAR}$. The experiments were carried out in two phases, Single Node performance and optimizations, and Multi-Node scaling. The feature size for $UoI_{LASSO}$ is kept a constant at 20,101 features across datasets to study the effect of communication. The number of samples are varied up to 51 million data points. For $UoI_{VAR}$, the dataset features range from 356 for a 128GB problem size to 1000 features for 8TB problem size and the number of samples are twice the size of the features.

\begin{table}[t]
\centering
\begin{center}
\begin{tabular}{ |c|c|c| } 
\hline
Performance Analysis& Data Size (GB) & \# of cores  \\ 
\hline
Single Node  & 16 & 68 \\
\hline
\multirow{7}{4em}{Weak Scaling} & 128 & 4,352\\
& 256 & 8,704\\
& 512 & 17,408\\
& 1024 & 34,816\\
&2048 & 69,632\\
& 4096 & 139,264\\
& 8192 &  278,528 \\
\hline
\multirow{3}{4em} {Strong Scaling} & \multirow{4}{2em}{1024} & 17,408 \\
& & 34,816\\
& & 69,632 \\
& &  139,264\\
\hline
 \end{tabular}
\caption{Performance Analysis setup for $UoI_{LASSO}$.}
\vspace{-0.5cm}
\label{experiments}
\end{center}
\end{table}

\begin{figure}[b]
\centering
\includegraphics[width=0.65\linewidth, scale=0.65]{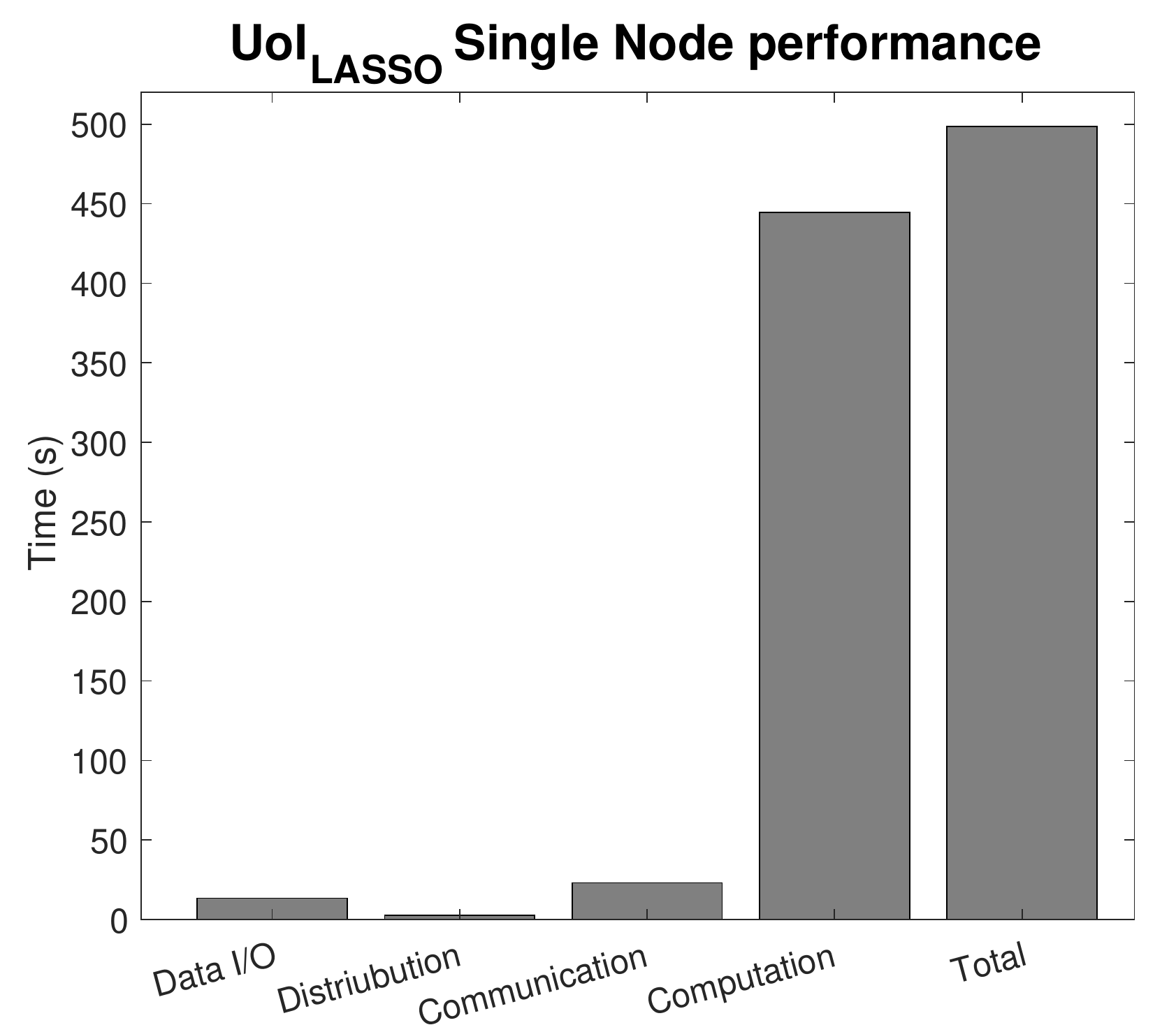}
\caption{$UoI_{LASSO}$ runtime number using Intel-MKL linear algebra library with $B_{1}=B_{2}=5$ and $q=8$.}
\vspace{-0.4cm}
\label{fig:single_lasso}
\end{figure}

\subsection{Performance and Scaling of $UoI_{LASSO}$}

\subsubsection{Single Node Performance}

The focus of single node performance analysis is to identify the potential bottlenecks in the program and optimize them. Post optimization, the performance improvement is calculated using a performance roofline model for both the program and the architecture (Xeon Phi) on which the program is executed. A $\approx16GB$ dataset with five selection and estimation bootstrap samples ($B_1 =  B_2 = 5$) and eight regularization parameters($q$) were chosen for single node optimization of the implementation.

Our initial analyses showed that the Matrix multiplication and Matrix-Vector product in LASSO-ADMM function were the bottlenecks. Execution of these operations accounted for almost 40.8\% of the total runtime and showed very poor performance with native Eigen3 library on Cori KNL. To alleviate the poor performance we implemented the \texttt{BLAS} operations for matrix multiply and matrix-vector product using the Intel-MKL library. Figure~\ref{fig:single_lasso} shows the runtime for single node run. Almost 90\% of the runtime is dominated by computation and less than 10\% by communication. All the \texttt{MPI} calls like \texttt{MPI\_Bcast}, \texttt{MPI\_Allreduce} etc., constitute the communication bar as shown in the Figure~\ref{fig:single_lasso}. More than 99\% of the communication time comes from \texttt{MPI\_Allreduce} call used to communicate the estimates by the distributed LASSO-ADMM function. MPI One-sided calls for distribution of the data is shown as `Distribution', while parallel-HDF5 data loading and saving is the `Data I/O' bar. We analyzed the program in detail with Intel Advisor~\cite{o2017intel} tool for the performance of various sections of the code.  The performance of matrix multiplication with Intel-MKL was 30.83 GFLOPS (Giga-Floating Point operations per second) with an arithmetic intensity (Floating point operations per byte of data moved from memory) of 3.59 FLOPs/Byte and the performance of matrix-vector multiplication was 1.12 GFLOPS with an arithmetic intensity of 0.32 FLOPs/Byte. Both the \texttt{BLAS} operations were found to be DRAM memory bound. The performance of the triangular solve function used by LASSO-ADMM function for matrix decomposition was 0.011 GFLOPS with an arithmetic intensity of 0.075 FLOPs/Byte.

\begin{figure}[t]
\centering
\includegraphics[width=0.85\linewidth, scale=0.85]{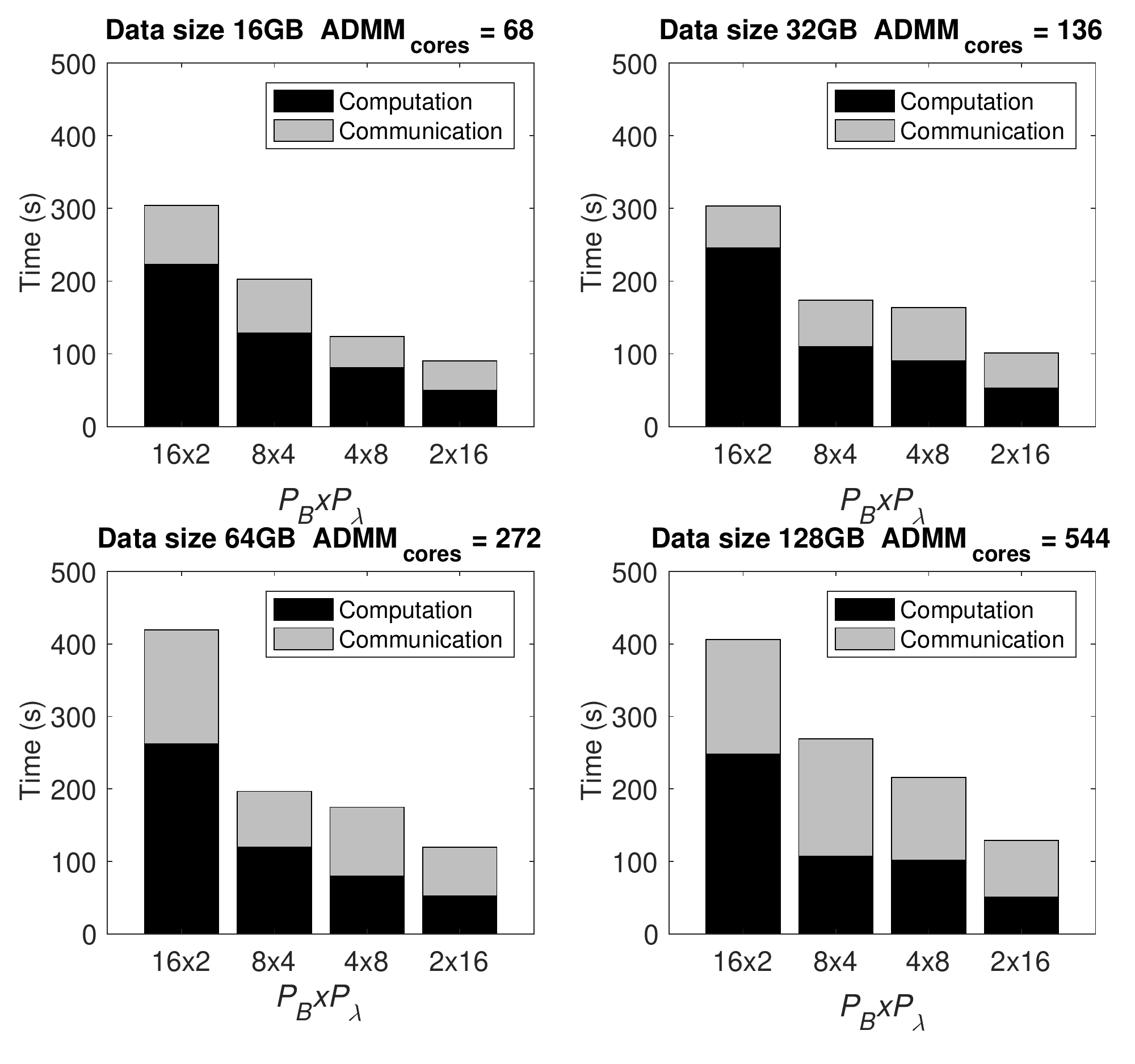}
\caption{Exploiting \textit{P$_B$} and \textit{P$_\lambda$} parallelism by increasing the dataset and ADMM$_{cores}$ by a factor of 2.}
\vspace{-0.4cm}
\label{fig:parallel_lasso}
\end{figure}

\subsubsection{Exploiting Algorithmic Parallelism}

The innate algorithmic parallelism exhibited by the $UoI_{LASSO}$ was exploited by having bootstrap level (\textit{P$_B$}), regularization parameter level (\textit{P$_\lambda$}) and ADMM computation level parallelism. These runs were performed on lower end of dataset spectrum, 16GB, 32GB, 64GB and 128GB with 2176, 4352, 8704 and 17,408 cores, respectively. The \textit{P$_B$} $\times$ \textit{P$_\lambda$} configuration used were 16$\times$2, 8$\times$4, 4$\times$8 and 2$\times$16 with $B_{1}=B_{2}=q=48$ for all the runs. The dataset size and the ADMM$_{cores}$ were doubled maintaining the parallelization configurations. The runtime of the different configurations are shown in Figure~\ref{fig:parallel_lasso}. Across various configurations the 2$\times$16 has a better runtime for all the datasets. Also across the dataset runs we can see a slight increase in the communication time for $ADMM_{cores} = 272$ and $ADMM_{cores} = 544$. This increase in the communication time was accounted by the \texttt{MPI\_Allreduce} call from LASSO-ADMM implementation to collectively converge at an estimate value.

\subsubsection{Multi-node Scaling}
The multi-node scaling analysis is carried out for weak scaling and strong scaling of $UoI_{LASSO}$ implementation. Parallel reading of the input file becomes an issue for multi-node scaling runs as 1000s of cores try to read the data in parallel. In an unoptimized run, the read time takes 10s of minutes which can worsen with an increase in the data size and the number of nodes.  For large datasets, the HDF5 input files are stripped into different Object Storage Targets (OSTs), explained in detail; in~\cite{howison2010tuning}. The files are stripped for 160 OSTs to achieve a faster reading time. reducing the read time of large datasets to a few seconds. The scaling runs were performed with no \textit{P$_B$} and \textit{P$_\lambda$} parallelism and dedicating all the cores to distributed LASSO-ADMM computation.



\begin{figure}[t]
\centering
\includegraphics[width=0.65\linewidth, scale=0.65]{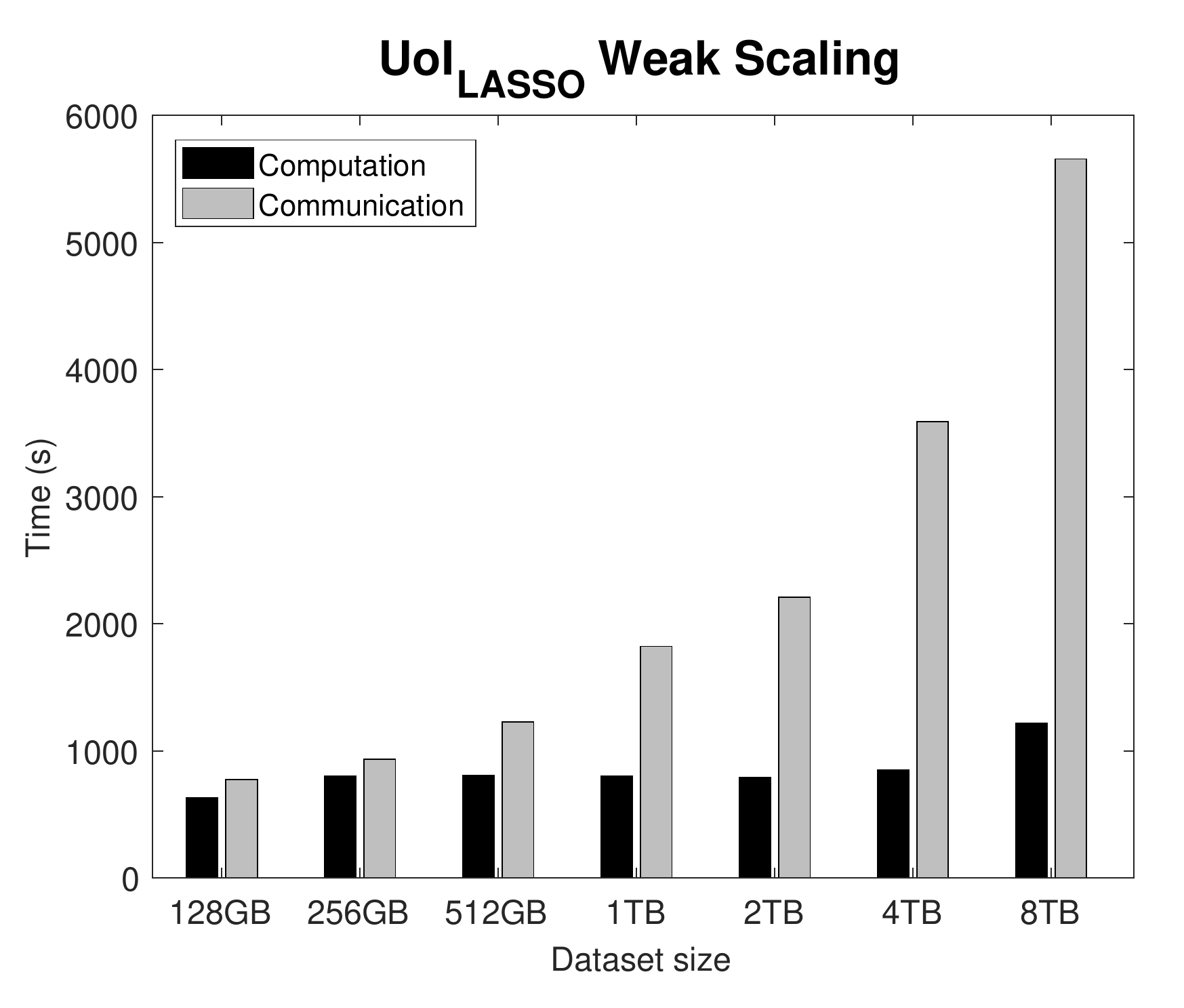}
\caption{Weak Scaling plot of $UoI_{LASSO}$}
\vspace{-0.4cm}
\label{fig:weak_lasso}
\end{figure}

\textbf{Weak Scaling:}
In weak scaling, the problem size associated with each compute core stays constant and additional computation cores are added when the size of the input dataset increases. We maintain a factor of 2 for our weak scaling runs, meaning as the dataset size is doubled the number of cores were also doubled (refer Table~\ref{experiments}). Figure~\ref{fig:weak_lasso} shows the weak scaling of $UoI_{LASSO}$. Since matrix multiplication contributes the most to the computation time, and since the problem size per compute core is almost the same across different configurations, we find that computation exhibits nearly ideal weak scaling with slight increase for 8TB. It is seen that \texttt{MPI\_Allreduce} call contributing to the communication time scales proportional to the increase in the core count. 

\begin{figure}[b]
\centering
\includegraphics[width=0.6\linewidth, scale=0.6]{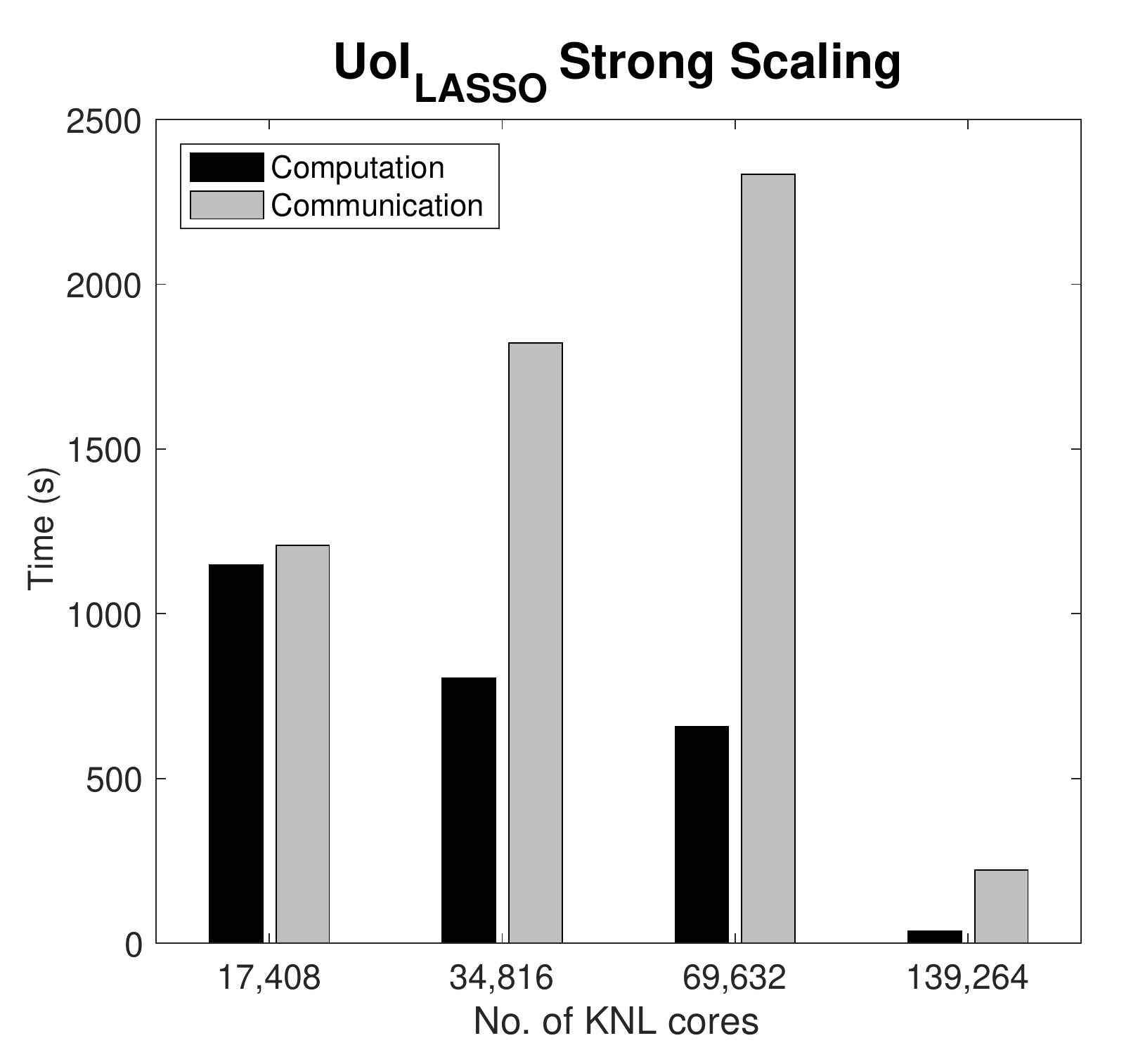}
\caption{Weak Scaling plot of $UoI_{LASSO}$ for 1TB dataset size.}
\vspace{-0.5cm}
\label{fig:strong_lasso}
\end{figure}

\textbf{Strong Scaling:}
In strong scaling, the problem size to be analyzed is kept as 1TB and the number of computation cores is increased from 17,408 to 139,264 (refer Table~\ref{experiments}). Figure~\ref{fig:strong_lasso} shows the results of the strong scaling run. The computation time shows a decreasing trend across different configurations due to the increase in the number of cores for the same dataset size. At 139,264 cores the computation goes below expected computation strong scaling trend, the reason being that the total size of the problem per core becomes small, which Intel-MKL library takes advantage of the AVX512 extensions making the matrix multiplication computed per core faster. Another reason for superlinear computation time is that the data matrix size per core becomes small which reduces the DRAM accesses. As seen in the weak scaling runs communication time increase with increasing number of cores, but beyond 69,632 cores the LASSO-ADMM converges faster making the communication time almost equal to the ideal strong scaling.  

\subsection{Performance and Scaling of $UoI_{VAR}$}

\begin{table}[t]
\centering
\begin{center}
\begin{tabular}{ |c|c|c| } 
\hline
Performance Analysis& Problem Size (GB) & \# of cores  \\ 
\hline
Single Node  & 16 & 68 \\
\hline
\multirow{7}{4em}{Weak Scaling} & 128 & 2,176\\
& 256 & 4,352\\
& 512 & 8,704\\
& 1024 & 17,408\\
&2048 & 34,816\\
& 4096 & 69,632\\
& 8192 &  139,264\\
\hline
\multirow{3}{4em} {Strong Scaling} & \multirow{4}{2em}{1024} & 4,352 \\
& & 8,704\\
& & 17,408 \\
& &  34,816\\
\hline
 \end{tabular}
\caption{Performance Analysis setup for $UoI_{VAR}$.}
\vspace{-0.6cm}
\label{tab:var}
\end{center}
\end{table}
  
\subsubsection{Single Node Performance} 
The Algorithm~\ref{algoVAR} creates a high dimensional matrix by Kronecker Product for each bootstrap subsample. The resultant matrix has a block diagonal structure with high sparsity. From Algorithm~\ref{algoVAR}, if the input data is dense the sparsity of the problem can be calculated as $1-\frac{1}{p}$, where $p$ is the number of features of the input dataset.


\begin{figure}[b]
\centering
\includegraphics[width=0.65\linewidth, scale=0.65]{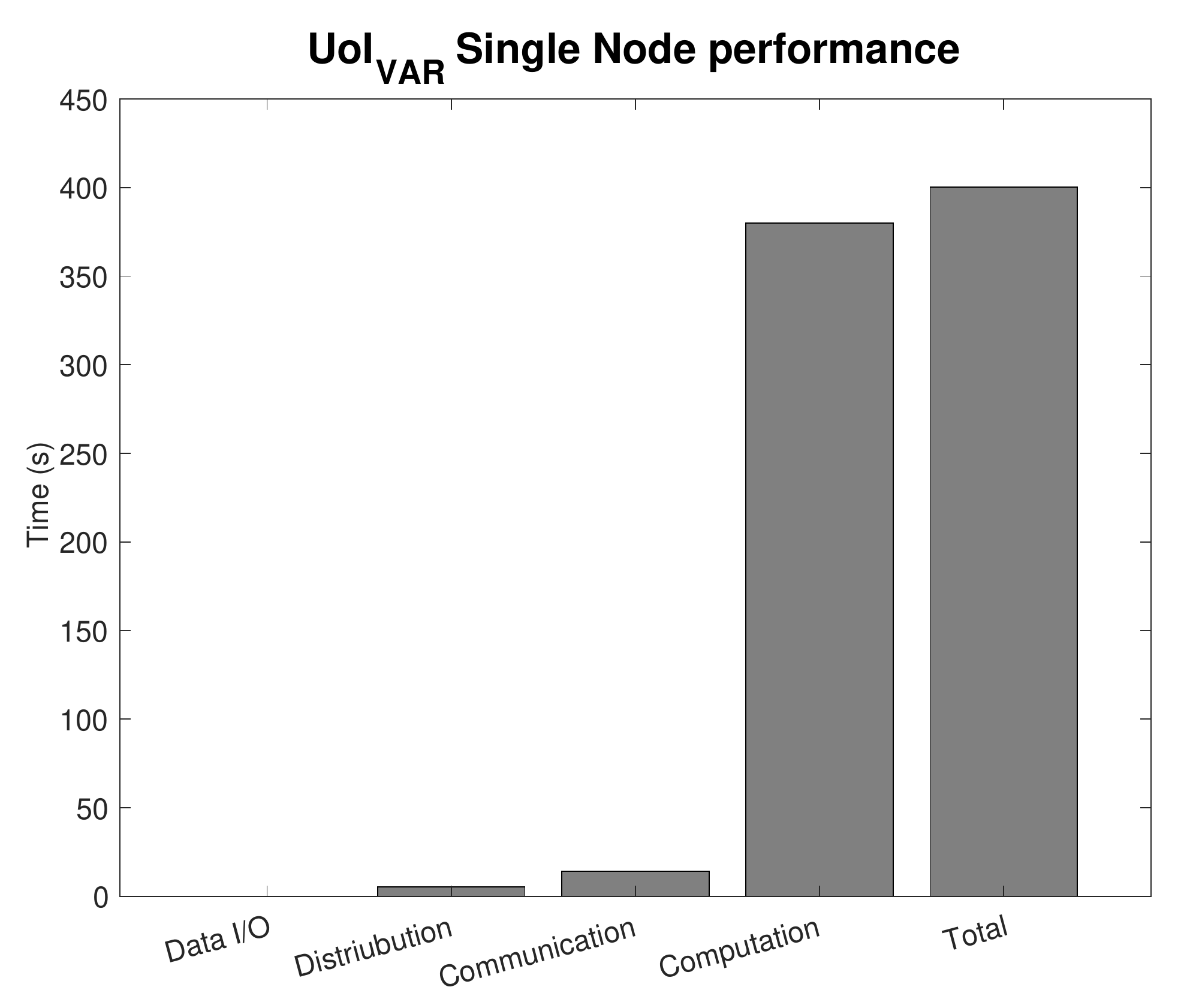}
\caption{$UoI_{VAR}$ single node runtime with Eigen Sparse C++ LASSO-ADMM implementation.}
\vspace{-0.4cm}
\label{fig:single_var}
\end{figure}

For example, if a dataset has 95 features, the resultant matrix post Kronecker Product has a sparsity of 98.94\%. So it is intuitive to exploit this sparsity by utilizing sparse linear algebra libraries. Figure~\ref{fig:single_var} shows the single node run of the $UoI_{VAR}$ implementation with Eigen3 Sparse C++ LASSO-ADMM. It is seen that the total execution time is dominated by the computation time. Like $UoI_{LASSO}$, communication is predominantly from LASSO-ADMM \texttt{MPI\_Allreduce} call. The distribution time denotes the distributed Kronecker Product and vectorization. $UoI_{VAR}$ implementation was also analyzed with the Intel Advisor software for performance metrics. The performance of sparse matrix multiplication was 1.08 GFLOPS with 0.15 arithmetic intensity and the performance of matrix-vector multiplication was 2.08 GFLOPS/sec with 0.33 arithmetic intensity. The performance of Simplicial Cholesky function and the triangular solve function used for matrix decomposition were 0.054 GFLOPS and 0.0082 GFLOPS respectively.


\begin{figure}[t]
\centering
\includegraphics[width=0.9\linewidth, scale=0.9]{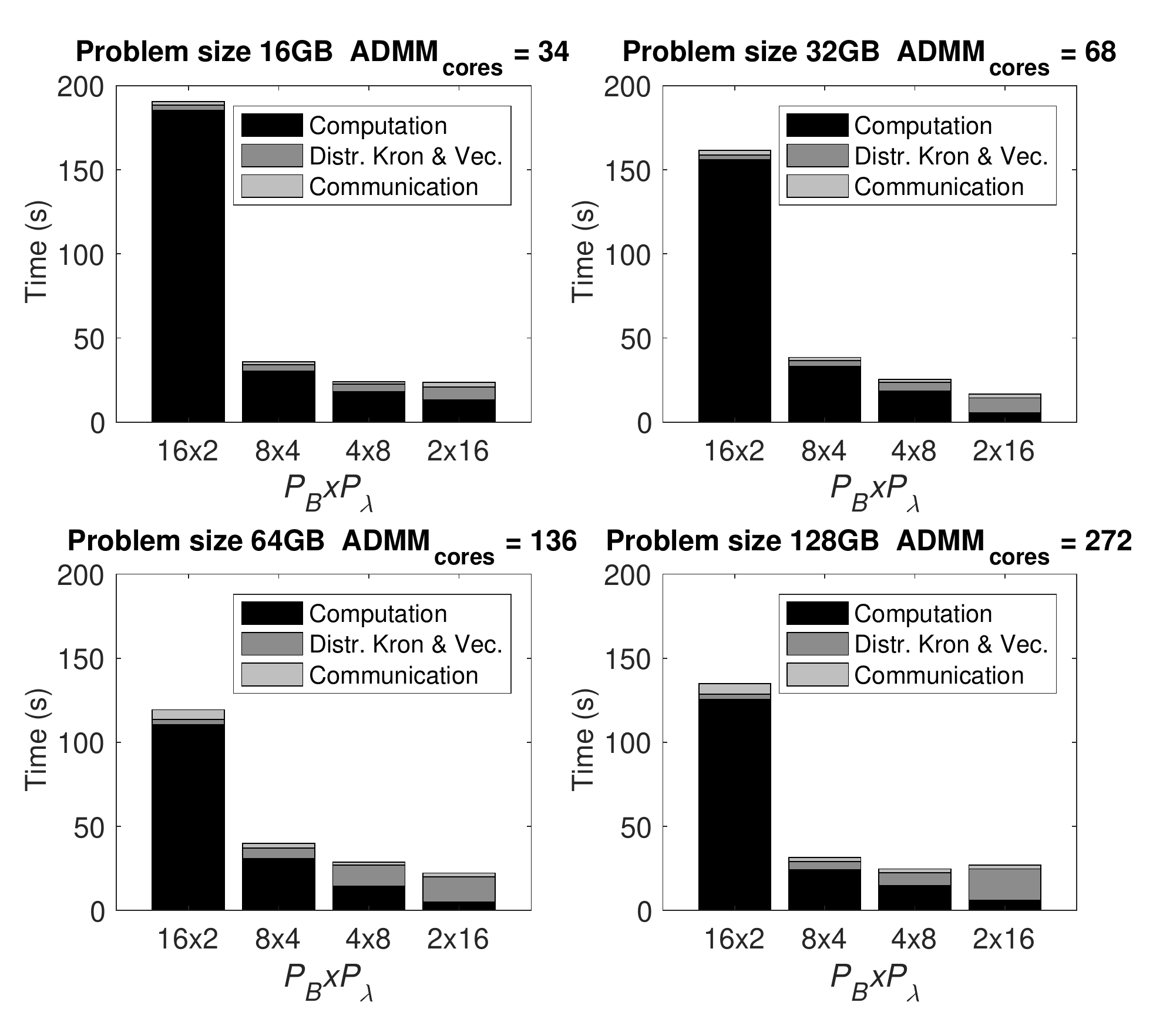}
\caption{Exploiting algorithmic parallelism of $UoI_{VAR}$.}
\vspace{-0.4cm}
\label{fig:parallel_var}
\end{figure}

\subsubsection{Exploiting Algorithmic Parallelism}
To exploit the algorithmic parallelism exhibited by UoI algorithms, a series of runs with \textit{P$_B$} and \textit{P$_{\lambda}$} parallelisms were performed. The runs were carried out for problem set sizes of 16GB, 32GB, 64GB and 128GB. The number of ADMM$_{cores}$ were doubled with doubling the problem size. The runs were performed for $B_{1}=B_{2}=32$ and $q=16$. The computation dominates the execution time, which decreases with increases in parallelism of \textit{P$_\lambda$} as shown in the Figure~\ref{fig:parallel_var}. It can also be noted that as the \textit{P$_\lambda$} parallelism increases the Kronecker Product of Vectorization time increases. From Algorithm~\ref{algoVAR} (lines 5, 21 and 22) the distributed Kronecker product and Vectorization is done for each bootstrap, and thus by reducing \texttt{P$_B$} parallelization increases the distribution time across different problem sets.

\subsubsection{Multi-node Scaling}
Weak and Strong scaling analyses were performed for $UoI_{VAR}$. The $UoI_{VAR}$ has a very different problem to $UoI_{LASSO}$ for data loading and distribution. The dataset size is very small compared to the problem size that is created during runtime. Distributed Kronecker Product and Vectorization creates this problem size which is of $O(p^3)$ ($p$ is the number of features) compared to the input data size. So unlike $UoI_{LASSO}$ distribution strategy only a few processes read the data in parallel and via MPI One-sided communication builds the problem.

\begin{figure}[t]
\centering
\includegraphics[width=0.65\linewidth, scale=0.65]{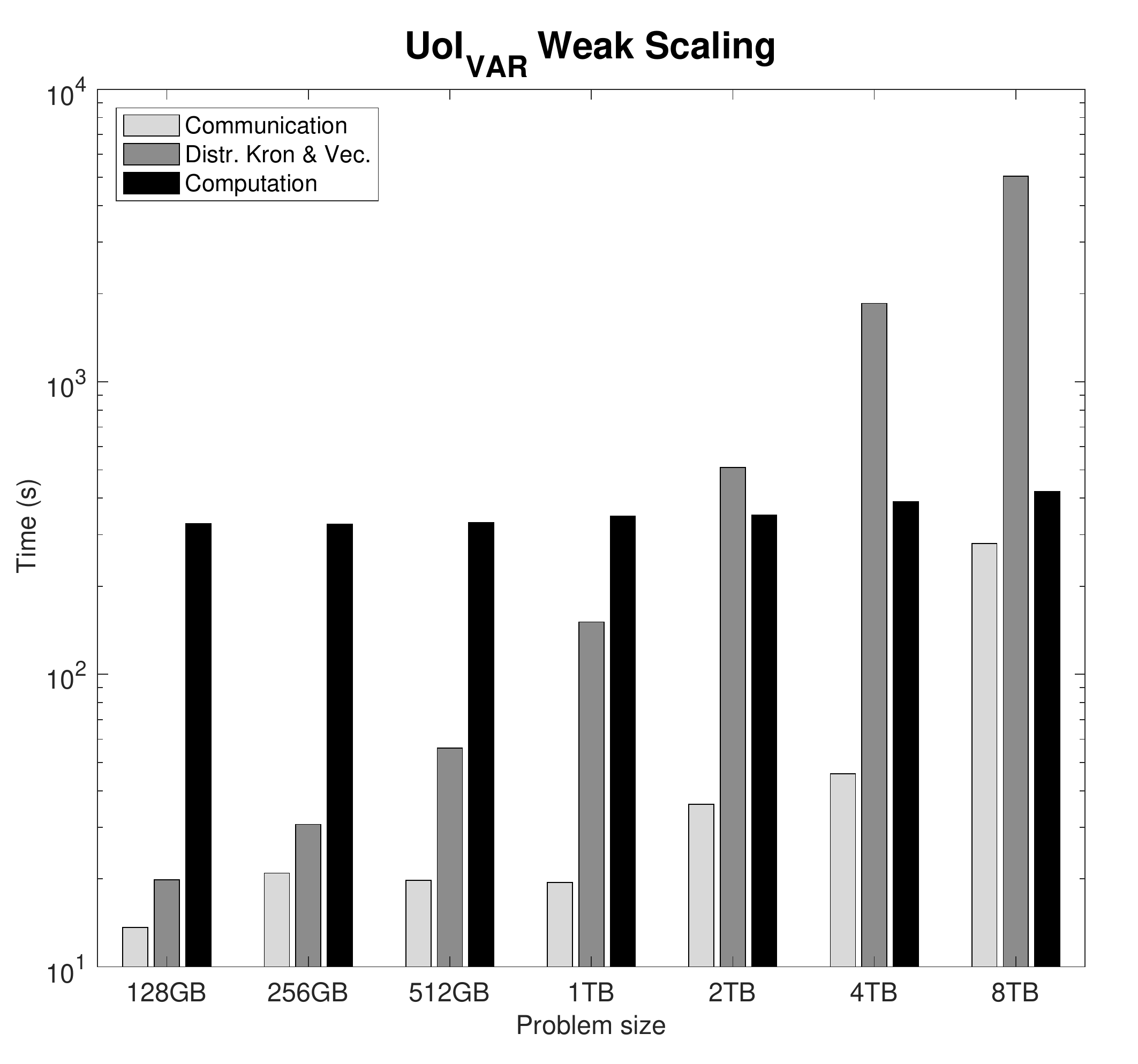}
\caption{Weak Scaling plot of $UoI_{VAR}$ in logarithmic scale.}
\vspace{-0.4cm}
\label{fig:weak_var}
\end{figure}

\textbf{Weak Scaling:} The weak scaling plot for $UoI_{VAR}$ is shown in the Figure~\ref{fig:weak_var} for $B_{1}=30, B_{2}=20, q=20$, with no \textit{P$_B$} or \textit{P$_\lambda$} parallelization. The Y-Axis in Figure~\ref{fig:weak_var} is given in a log-scale to show logarithmic increase in the distribution time. It can be seen that computation has almost ideal weak scaling, and the communication time also increases with increase in core count as seen in $UoI_{LASSO}$. The distributed Kronecker Product and Vectorization do not scale well and is proportional to the increase in the cores and problem size. One of the main reasons for this trend is the cubical trend of the problem size explosion ($O(p^3)$, where $p$ is the number of features). Since the input data is read only by a few processes ($n\_readers$) and all other processes repeatedly access $n\_readers$ cores to build the problem via \texttt{MPI\_Get}, there is a bottleneck. One way to mitigate the problem is by \textit{P$_B$} parallelization as seen in Figure~\ref{fig:parallel_var}.

\begin{figure}[b]
\centering
\includegraphics[width=0.65\linewidth, scale=0.65]{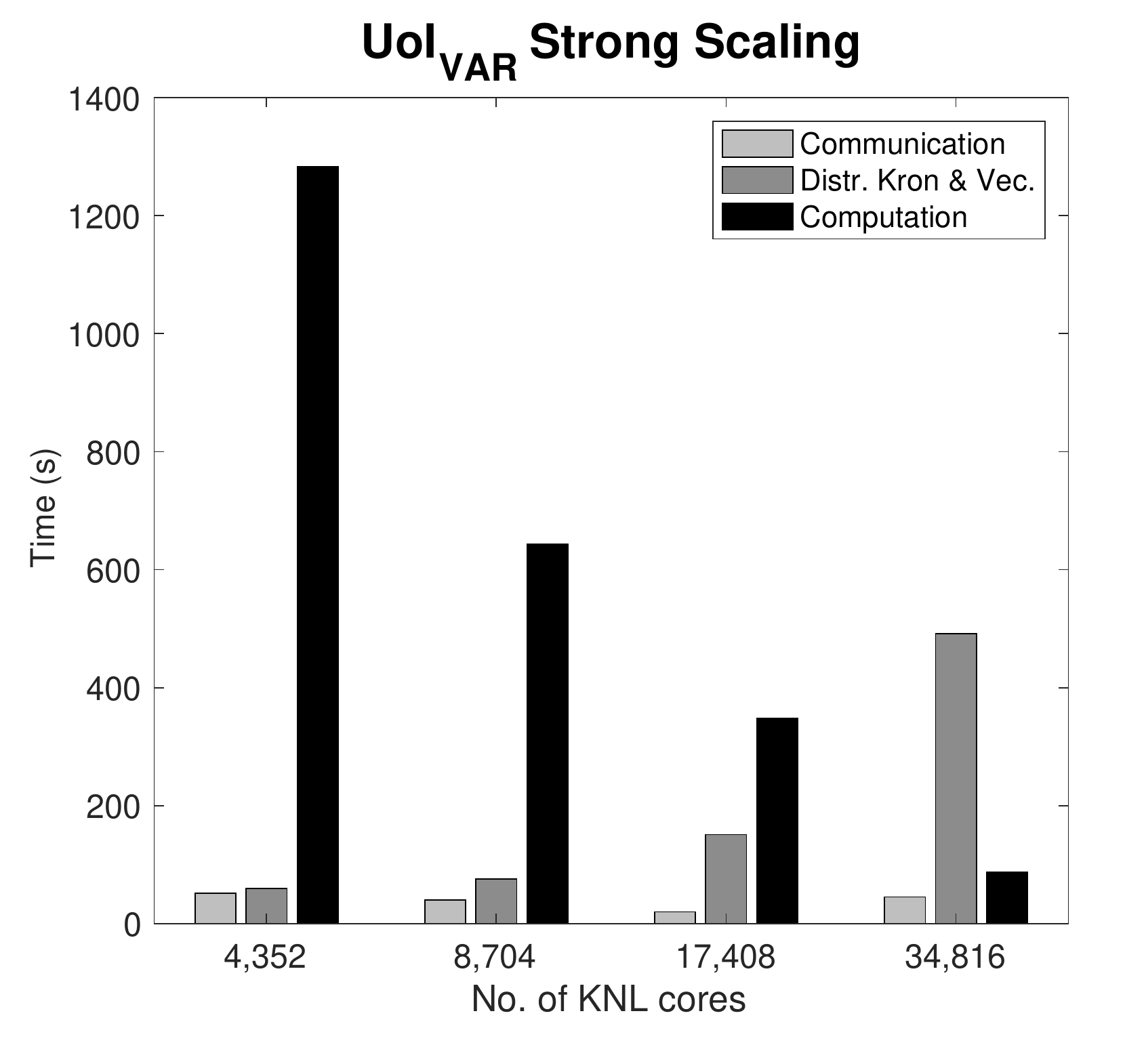}
\caption{Strong Scaling plot of $UoI_{VAR}$ for 1TB problems size.}
\vspace{-0.4cm}
\label{fig:strong_var}
\end{figure}

\textbf{Strong Scaling:} The strong scaling plot for $UoI_{VAR}$ is shown in the Figure~\ref{fig:strong_var}. It can be seen that across increasing core sizes computation time has an almost ideal strong scaling. The reason for an ideal computation time is because of the matrix-vector multiplication which has a linear scaling, and as discussed earlier, the sparse matrix product is done using Sparse Eigen C++. Even though the communication does not have an ideal scaling it minimally affects the total runtime of the program. The distributed Kronecker Product and Vectorization scales proportional to the increase in the number of cores like the weak scaling.

\section{Discussion}
There is a trade-off between communication and computation in $UoI_{LASSO}$, as shown in Figure~\ref{fig:weak_lasso}. When the data size per core increases the computation time increases because the computation bottlenecks are \texttt{BLAS} \texttt{gemm} and \texttt{gemv} operations. On the other hand for large datasets, the runtime of the code is determined by communication via \texttt{MPI\_Allreduce} call, whereas the computation has a near ideal scaling. Almost 98\% of the communication time, \texttt{MPI\_Allreduce}, that is seen in weak and strong scaling is from `Model Selection' section of the algorithm. To mitigate this effect of communication on the runtime, \textit{P$_B$} and  \textit{P$_\lambda$} parallelism can be adopted as shown in Figure~\ref{fig:parallel_lasso} based on availability of resources. To improve the communication bottleneck, especially \texttt{MPI\_Allreduce}, different optimization strategies explained in~\cite{thakur2005optimization} can be incorporated.  

As shown in Figure~\ref{fig:weak_var}, there is a trade-off between computation and distribution in $UoI_{VAR}$. For smaller problem sizes computation dominates the program runtime and for larger problem sizes (especially for problem sizes 2TB and above) distribution dominates the total program runtime. The reason for this being the problem size explosion, where for a small input data size the distributed Kronecker Product and Vectorization creates a large matrix. One of the ways to avoid the problem is by utilizing \textit{P$_B$} parallelism. Another way to alleviate this issue is by using communication avoiding algorithms and using local computation modules to create the matrix and then have a one-time communication to create the large matrix. As shown in Figure~\ref{fig:strong_var}, it can be seen that the distribution is directly proportional to the increase in the number of cores. 

\section{Real Dataset}

A single session non-human primate reaching task dataset \cite{odoherty583331} was analyzed using $UoI_{VAR}$. Monkey reaching behavioral tasks were recorded in \cite{odoherty583331} with two monkey subjects. Some of the recorded datasets consist of spikes for both the motor cortex (M1) and, the somatosensory cortex (S1) recordings for 192 electrodes as features.  The recorded spikes had 51111 samples recorded for one session. In the VAR model, the dataset created a problem size $\approx$  1.3TB. The problem was executed on 81,600 cores on Cori KNL. The computation and communication times were found to be 96.9s and 1598.72s, respectively. The distribution time was 3034.4s.

\section{Conclusion}
In this paper, we have implemented $UoI_{LASSO}$ and $UoI_{VAR}$ using the Union of Intersection framework which promises high interpretability with increased prediction. We have presented a scalable $UoI_{LASSO}$ and $UoI_{VAR}$ implementation with a randomized data distribution strategy for HDF5 to aid parallel bootstrap subsampling for $UoI_{LASSO}$ and distributed Kronecker product and vectorization for $UoI_{VAR}$. The single-node performance evaluation and multi-node scaling runs used a wide range of sizes of synthetic datasets. Our weak and strong scaling analyses show that $UoI_{LASSO}$ is communication bound and $UoI_{VAR}$ is distribution bound for large datasets and problem sizes, respectively.  Finally, we have presented discussions on the two algorithms implemented and possible future directions to improve the algorithm for better communication scaling. With our  $UoI_{VAR}$ implementation, we have created the largest VAR model (1,000 nodes) we are aware of.


\bibliographystyle{IEEEtrans}
\bibliography{bare_conf}

\begin{thebibliography}{10}

\bibitem{national2013frontiers}
National~Research Council et~al.
\newblock {\em Frontiers in massive data analysis}.
\newblock National Academies Press, 2013.

\bibitem{8352088}
Kristofer~E. Bouchard et~al.
\newblock International neuroscience initiatives through the lens of
  high-performance computing.
\newblock {\em Computer}, 51(4):50--59, April 2018.

\bibitem{bouchard2017union}
Kristofer~E. Bouchard et~al.
\newblock {Union of Intersections (UoI) for Interpretable Data Driven Discovery
  and Prediction}.
\newblock In {\em Advances in Neural Information Processing Systems}, pages
  1078--1086, 2017.

\bibitem{folk2011overview}
Mike Folk et~al.
\newblock {An overview of the HDF5 technology suite and its applications}.
\newblock In {\em Proceedings of the EDBT/ICDT 2011 Workshop on Array
  Databases}, pages 36--47. ACM, 2011.

\bibitem{eigenweb}
Ga\"{e}l Guennebaud, Beno\^{i}t Jacob, et~al.
\newblock {Eigen v3}.
\newblock http://eigen.tuxfamily.org, 2010.

\bibitem{wang2014intel}
Endong Wang et~al.
\newblock {Intel Math Kernel Library}.
\newblock In {\em High-Performance Computing on the Intel Xeon Phi™|}, pages
  167--188. Springer, 2014.

\bibitem{o2017intel}
K~O’Leary, I~Gazizov, A~Shinsel, R~Belenov, Z~Matveev, and D~Petunin.
\newblock {Intel Advisor Roofline Analysis: A New Way to Visualize Performance
  Optimization Trade-offs}.
\newblock {\em Intel Software: The Parallel Universe}, 27:58--73, 2017.

\bibitem{howison2010tuning}
Mark Howison et~al.
\newblock {Tuning HDF5 for lustre file systems}.
\newblock Technical report, Ernest Orlando Lawrence Berkeley National
  Laboratory, Berkeley, CA (US), 2010.

\bibitem{thakur2005optimization}
Rajeev Thakur et~al.
\newblock {Optimization of collective communication operations in MPICH}.
\newblock {\em The International Journal of High Performance Computing
  Applications}, 19(1):49--66, 2005.

\bibitem{odoherty583331}
Joseph~E. O'Doherty et~al.
\newblock {Nonhuman Primate Reaching with Multichannel Sensorimotor Cortex
  Electrophysiology}.
\newblock \url{https://doi.org/10.5281/zenodo.583331}.

\end{thebibliography}

\end{document}